\documentclass[US, 10pt, conference]{ieeeconf}    
\IEEEoverridecommandlockouts
\usepackage{amsmath,amsfonts}
\usepackage{algorithm}
\usepackage{algpseudocode}
\usepackage{array}
\usepackage{textcomp}
\usepackage{stfloats}
\usepackage{url}
\usepackage{verbatim}
\usepackage[pdftex]{graphicx}
\usepackage{cite}
\usepackage{wrapfig}
\usepackage{subcaption}
\usepackage{xcolor}
\usepackage{tabularx}
\usepackage{booktabs} 
\usepackage{multirow} 
\usepackage{soul}
\usepackage{amssymb}
\usepackage{threeparttable}

\begin{document}

\title{SpikingSoft: A Spiking Neuron Controller \\ for Bio-inspired Locomotion with Soft Snake Robots
}

\author{Chuhan Zhang, Cong Wang, Wei Pan, Cosimo Della Santina
\thanks{This work is supported by the
European Union’s Horizon Europe Program from Project EMERGE - Grant Agreement No.101070918 and the TU Delft AI Labs programme.}
\thanks{Chuhan Zhang, Cong Wang, Cosimo Della Santina are with the Department of Cognitive Robotics, Faculty of Mechanical Engineering, Delft University of Technology, Delft, Netherlands (e-mail: \{C.Zhang-8, C.Wang-17, C.DellaSantina\}@tudelft.nl).}
\thanks{Wei Pan is with the Department of Computer Science, The University of Manchester, Manchester, United Kingdom (e-mail: wei.pan@manchester.ac.uk).}
}

\maketitle

\begin{abstract}
Inspired by the dynamic coupling of moto-neurons and physical elasticity in animals, this work explores the possibility of generating locomotion gaits by utilizing physical oscillations in a soft snake by means of a low-level spiking neural mechanism.
To achieve this goal, we introduce the Double Threshold Spiking neuron model with adjustable thresholds to generate varied output patterns. This neuron model can excite the natural dynamics of soft robotic snakes, and it enables distinct movements, such as turning or moving forward, by simply altering the neural thresholds.
Finally, we demonstrate that our approach, termed SpikingSoft, naturally pairs and integrates with reinforcement learning. The high-level agent only needs to adjust the two thresholds to generate complex movement patterns, thus strongly simplifying the learning of reactive locomotion. Simulation results demonstrate that the proposed architecture significantly enhances the performance of the soft snake robot, enabling it to achieve target objectives with a 21.6\% increase in success rate, a 29\% reduction in time to reach the target, and smoother movements compared to the vanilla reinforcement learning controllers or Central Pattern Generator controller acting in torque space.

\end{abstract}

\begin{figure*}[h]
    \centering
    \begin{minipage}[b]{0.6\linewidth}
        \begin{subfigure}[b]{0.39\textwidth}
            \centering
            \includegraphics[width=\linewidth]{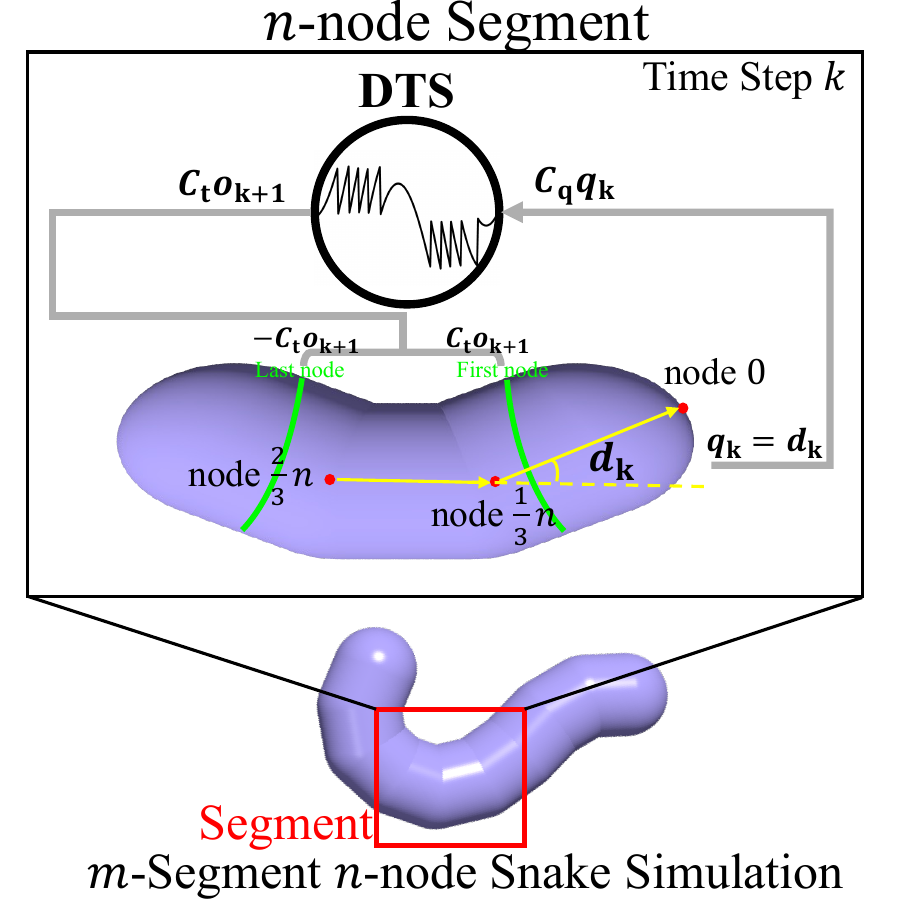}
        \end{subfigure}%
        \begin{subfigure}[b]{0.6\textwidth}
            \centering
            \includegraphics[width=\linewidth]{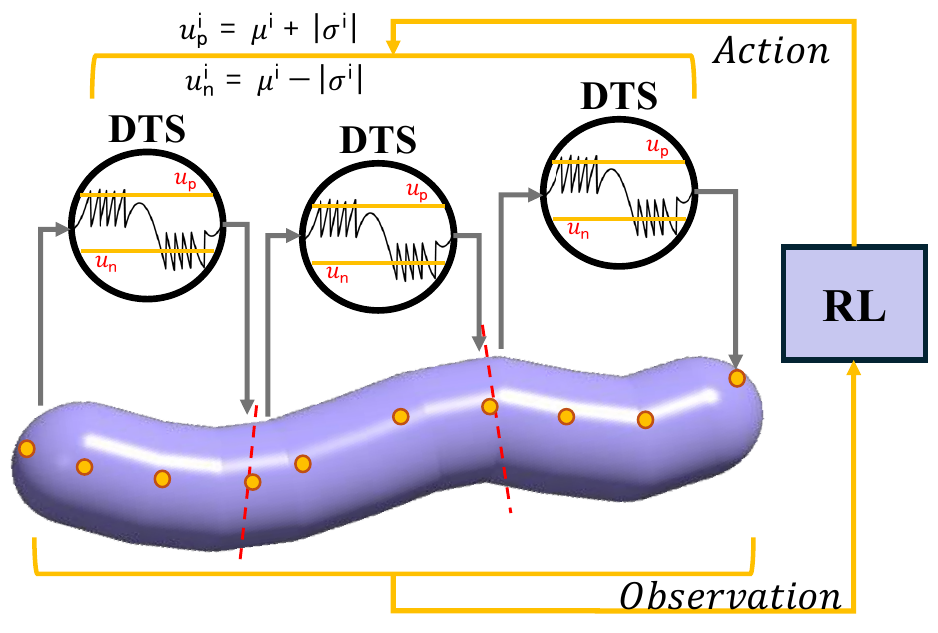}
        \end{subfigure}
    \end{minipage}
    \begin{minipage}[b]{0.39\linewidth}
    \caption{The pipeline of the proposed biologically plausible control architecture. Left: The SpikingSoft controller structure. The oscillatory behavior $d$ is the input of the proposed spiking neuron. $C_{\mathrm{t}}$ and $C_{\mathrm{q}}$ are hyperparameters to adjust the scale. $o$ is the output spikes. Right: The reinforcement learning framework of a 3-segment 3-node snake with SpikingSoft controllers. $\mu^{\mathrm{i}}$ and $\sigma^{\mathrm{i}}$ are taken from the RL actions, composing the double thresholds $u_{\mathrm{n}}^{\mathrm{i}}$ and $u_{\mathrm{p}}^{\mathrm{i}}$ for the controller in the $i^{\mathrm{th}}$ segment.
    }
    \label{fig:overall}
    \end{minipage}
    \vspace{-0.5cm}
\end{figure*}

\section{Introduction}
Animals exhibit remarkable locomotion capabilities, as they are able to generate complex and adaptive gaits to traverse diverse terrains and perform complex tasks. This natural ability to coordinate movement has inspired roboticists to design gait controllers \cite{kuo2023deep, khan2023review}. In this work, we specifically focus as a case study on an especially complex and simultaneously extremely relevant locomotion scenario: soft robotic snake slithering.

With their ability to move within confined and unstructured spaces, soft snake robots show great value in fields ranging from medical procedures to search and rescue operations \cite{qi2020novel,seetohul2022snake,liu2021review,della2020soft,rajashekhar2024developments}. 
Traditionally, many works have employed kinematic-based controllers to control the movement of rigid snake robots. The serpenoid model \cite{ma1999analysis}, which draws inspiration from the natural udulatory motion of snakes and utilizes sinusoidal wave functions to describe the body shape during locomotion, is capable of generating various gaits to effectively control the snake robots' movement  \cite{li2021adaptive, arachchige2021soft}. \cite{zhao2021multigait} proposed five kinematic models representing different gaits for snake robots, which generate target motor angles that drive the locomotion task. Central Pattern Generators (CPGs), which replicate the rhythmic movements observed in animals, offer another robust method for gait generation \cite{ijspeert2008central}. The oscillatory outputs of CPGs can be employed to generate desired joint angles \cite{wu2010cpg}. 
To adapt to more complex scenarios, such as reaching random targets or uneven terrains, kinematic-based controllers are often combined with Reinforcement Learning (RL) frameworks. These hybrid approaches leverage the strengths of both methods to improve adaptability and performance. For instance, \cite{jia2021coach} proposed an RL framework to learn the amplitude and offset increments, which enhance the performance of serpenoid controllers on rigid wheeled snake robots. Similarly, \cite{liu2023energy, bing2022simulation} introduce RL frameworks designed to learn the joint angles within the serpenoid curve. In addition, \cite{shi2020deep} applies an RL framework to learn joint angle velocities, enabling the generation of diverse gaits for the robot. Furthermore, \cite{qiu2021reinforcement} explores the use of RL to learn parameters that adjust the serpenoid trajectories within a CPG model.
Kinematic-based controllers, while effective for rigid snake robots, often face significant challenges when applied to soft snake robots. This limitation arises from the difficulty in accurately tracking the desired joint positions or trajectories due to the soft body’s nonlinear and elastic mechanics. As a result, model-free controllers together with RL frameworks have been increasingly adopted for controlling soft snake robots. \cite{liu2023reinforcement} utilizes RL frameworks to train CPG controllers to directly produce motor commands, enabling the generation of diverse gaits without relying on predefined models. Learning CPG by using RL also gains success for locomotion tasks in other various robotic systems \cite{bellegarda2022cpg, shafiee2024manyquadrupeds}. However, despite the effectiveness of CPG controllers in generating continuous control commands, they offer limited exploration of the self-oscillatory behaviors of soft bodies, restricting efficient control. To address this, our research proposes an event-triggered controller that can excite and utilize the natural oscillations of the soft material, improving control performance and adaptability.

Inspired by the natural example\cite{bello2024motor}, we consider in this work the use of more biologically plausible spiking neurons as model-free gait controllers. Spiking neurons are widely
recognized as more biologically realistic representations of
neuronal behavior than traditional artificial neurons\cite{schuman2022opportunities, abadia2021cerebellar}.
Unlike traditional neurons that produce continuous values,
spiking neurons employ discrete events or \textit{spikes}. Incorporating the complex temporal change of membrane potentials, spiking neurons offer richer and more detailed neuronal dynamics than traditional neurons with a simplified and static nature, which makes them fit for combining with soft robot control. 
Today, many works use SNN for traditional control tasks such as OpenAI Gym tasks \cite{brockman2016openai}. Many works combine SNN into RL frameworks to learn the Q value \cite{chen2022deep, akl2021porting} or the actor networks \cite{tang2021deep,chen2024fully}. Some other work has proposed RL strategies based on SNN learning rules \cite{bing2018end, liu2023spiking}. SNN can also be used for perception in the control pipeline \cite{jiang2020target}. However, the integration of spiking neurons as controllers into soft snake robots is still in its early stages of exploration. Also, the binary output of spiking neurons—limited to zeros and ones—restricts their use in complex control systems requiring versatile motor commands. To overcome this, research has shifted towards spiking neurons with variable thresholds, improving their application in computer vision and data processing \cite{ding2022biologically, zhang2023low}. \cite{yu2021constructing} demonstrated success in image recognition using a dual-threshold neuron and ANN-SNN conversion.

To conclude, this work explores the potential of spiking neuron models as the gait generator in soft robots for effective snake locomotion, where neuron activities such as accumulation, firing, and resetting synchronize with robot movement in a closed-loop manner. To bridge gaps in applying dynamic spiking properties to soft robot elasticity, we propose a single-neuron-based controller that adjusts spiking patterns based on oscillation signals without a predefined model. Using standard RL methods, we demonstrated through simulations that this controller enables our soft snake robot to achieve bionic gait locomotion. 
Note that spiking neural networks have already been explored in the context of robotic locomotion with very limited DOF robot segments where the spikes function as simple on/off switches \cite{rostro2015cpg,tieck2019combining,wang2019locomotion,lele2020learning,lele2021end,yamazaki2022spiking,jiang2023fully}.
The spiking mechanism has never been used to generate high-frequency outputs for controlling a high DOF soft body, particularly by harnessing the internal oscillatory dynamics of the soft structure, which is instead the focus of our work. Also, all these works do not focus on snake systems, but on legged robots. Interestingly, a few works \cite{chen2017toward,jiang2020target} have looked into the combination of SNNs and snake locomotion, but with different strategies and - again - not in the context of generating gaits as a direct model-free controller.

{The contributions of this paper can be summarized as follows}: \textbf{(1)} We are the first to address the integration of bio-inspired spiking neurons as gait controllers in the control architecture of bionic soft robots. \textbf{(2)} We propose a novel Double Threshold Spiking neuron (DTS) model, based on which we propose a SpikingSoft controller for the soft snake robot. The SpikingSoft controller can generate rich output spike patterns by simply tuning the thresholds to control the soft snake robot moving freely, smoothly and stably in a model-free manner. \textbf{(3)} With the proposed controller, we significantly improve the success rate of the soft snake robot in the target-reaching task, allowing it to finish efficiently in a bionic gesture.

The remainder of the paper is organized as follows. Section \ref{methods} introduces our proposed DTS neuron and controller. Section \ref{experiments} gives the details and results of the experiment. Section \ref{conclusion} concludes the work and discusses the limitation.

\section{Methods}
\label{methods}

In this section, the proposed DTS neuron model is illustrated in Section \ref{methods_sn}. Finally, the SpikingSoft controller for the soft snake robot is described in Section \ref{methods_contoller}.

\subsection{Double Threshold Spiking neuron}
\label{methods_sn}

In the SNN community, the Leaky Integrate-and-Fire (LIF) neuron model is widely used. In the LIF model, the membrane potential $u$ accumulates over time and returns to the reset position when it reaches the threshold, and the neuron emits a spike. However, binary spikes consisting solely of 0s and 1s present challenges in mapping various motor commands. This constraint makes it difficult to perform complex control tasks on soft robots. To address this limit, we propose a Double Threshold Spiking neuron (DTS) model, which can be described using these equations:
\begin{figure}[h]
\begin{minipage}[t]{0.4\linewidth}
\begin{equation}
\begin{aligned}
& \text{Membrane potential:} \\
&  \left\{ 
    \begin{aligned}
    & \tau \frac{\mathrm{d}u(t)}{\mathrm{d}t} = -u(t) + C_{\mathrm{q}}q(t), \\
    & \hspace{1.1cm} \text{if } u_{\mathrm{n}} \leq u(t) \leq u_{\mathrm{p}}, \\
    & \lim\limits_{\Delta \to 0^{+}}u({t}+\Delta) = 0, \text{else}
    \end{aligned}\right.
\end{aligned}
\nonumber
\end{equation}
\end{minipage}
\hspace{1.0cm} 
\begin{minipage}[t]{0.4\linewidth}
\begin{equation}
\begin{aligned}
& \text{Spike generation:} \\
& \Gamma(t) = C_{\mathrm{t}} o(t) = C_{\mathrm{t}} g(u(t)), \\
& g(u) = \left\{ 
      \begin{array}{ll}
        1, & \text{if } u < u_{\mathrm{n}}, \\
        -1, & \text{if } u > u_{\mathrm{p}}, \\
        0, & \text{else,}
      \end{array} 
    \right.
\end{aligned}
\label{formula_dts_continuous_1}
\nonumber
\end{equation}
\end{minipage}
\end{figure}
where $\tau \in \mathbb{R}$ is the membrane time constant and $u(t) \in \mathbb{R}$ represents the membrane potential of the neuron in continuous time $t$. $C_{\mathbf{q}} \in \mathbb{R}$ and $C_{\mathbf{t}} \in \mathbb{R}$ are constants for the input signal $q(t) \in \mathbb{R}$ and output torque $\Gamma(t) \in \mathbb{R}$ respectively. $u_{\mathrm{n}}, u_{\mathrm{p}} \in \mathbb{R}$ are two thresholds that satisfy condition $u_{\mathrm{n}} < u_{\mathrm{p}}$. Specifically, the neuron generates a negative spike of $o(t)=-1$ when the membrane potential exceeds the positive threshold $u_{\mathrm{p}}$. In contrast, when the membrane potential drops below the negative threshold $u_{\mathrm{n}}$, a positive spike of $o(t)=1$ is generated. No spike is generated if neither condition is satisfied and $o(t)=0$. The function $g(\cdot)$ is the Heaviside step function, which illustrates the spike fire process. Based on \eqref{formula_dts_continuous_1}, the discrete-time dynamics of the DTS neuron are expressed as:
\begin{equation}
\label{formula_dts_discrete}
\begin{aligned}
u^{\mathrm{k+1}} & = (1- |o^{\mathrm{k}}|)(1-\frac{\mathrm{d}t}{\tau})u^{\mathrm{k}}+\frac{\mathrm{d}t}{\tau} C_{\mathrm{q}} q^\mathrm{k+1}, \\
o^{k} & = \left\{ 
    \begin{array}{ll}
        1, & \text{if } u^{\mathrm{k}} < u_{\mathrm{n}},\\
        -1, & \text{if } u^{\mathrm{k}} > u_{\mathrm{p}},\\
        0, & \text{else.} \\
    \end{array}
    \right.  \Gamma^{\mathrm{k}} = C_{\mathrm{t}} o^{\mathrm{k}},
\end{aligned}
\end{equation}
where $k$ represents the discrete time step.

\begin{figure*}[ht]
\vspace{0.2cm}
    \centering
    \begin{minipage}[b]{0.71\linewidth}
        \includegraphics[width=\linewidth]{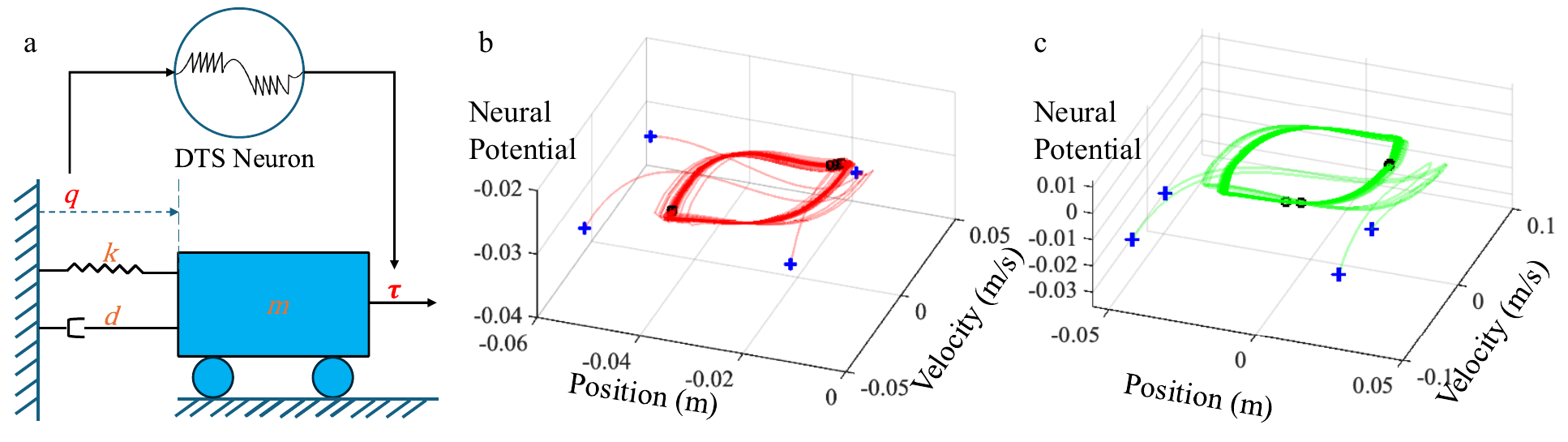}
    \end{minipage}
    \begin{minipage}[b]{0.27\linewidth}
    \caption{The phase space limit circle of the mass-spring-damper system. \textbf{a:} The mass-spring-damper system. Others: Three oscillation patterns controlled by a DTS neuron. \textbf{b:} $u_{\mathrm{n}}=-0.1, u_{\mathrm{p}}=-0.025$. \textbf{c:} $u_{\mathrm{n}}=-0.0025, u_{\mathrm{p}}=0.0025$.}
    \label{fig:car_system}
    \end{minipage}
\end{figure*}

Unlike traditional LIF neurons \cite{wu2018spatio}, DTS neurons only use positive and negative thresholds as variable parameters while ignoring classic input weights. Instead, a constant $C_{\mathrm{q}}$ is employed for the input signal $q$. This approach ensures the stability of the membrane potential in the accumulation phase.
In addition, the resting membrane potential is set at 0 in DTS neurons. This design facilitates the symmetric behavior of the neuron.
When the input signal is exactly opposite, such as $\sin$ and $-\sin$, the DTS neuron can generate opposite outputs.
Furthermore, unlike the dual threshold neurons introduced by \cite{yu2021constructing}, the positive threshold $u_{\mathrm{p}}$ and negative threshold $u_{\mathrm{n}}$ of DTS neurons are only relative in magnitude. These thresholds can simultaneously exceed or fall below zero. When $u_{\mathrm{p}}$ and $u_{\mathrm{n}}$ are both positive or negative, DTS neurons continuously generate spikes regardless of the input signal. This behavior occurs because the membrane potential for resetting is set to zero, which facilitates the neuron's ability to generate spikes at a high rate.

\begin{figure}[ht]
    \centering
    \includegraphics[width=\linewidth]{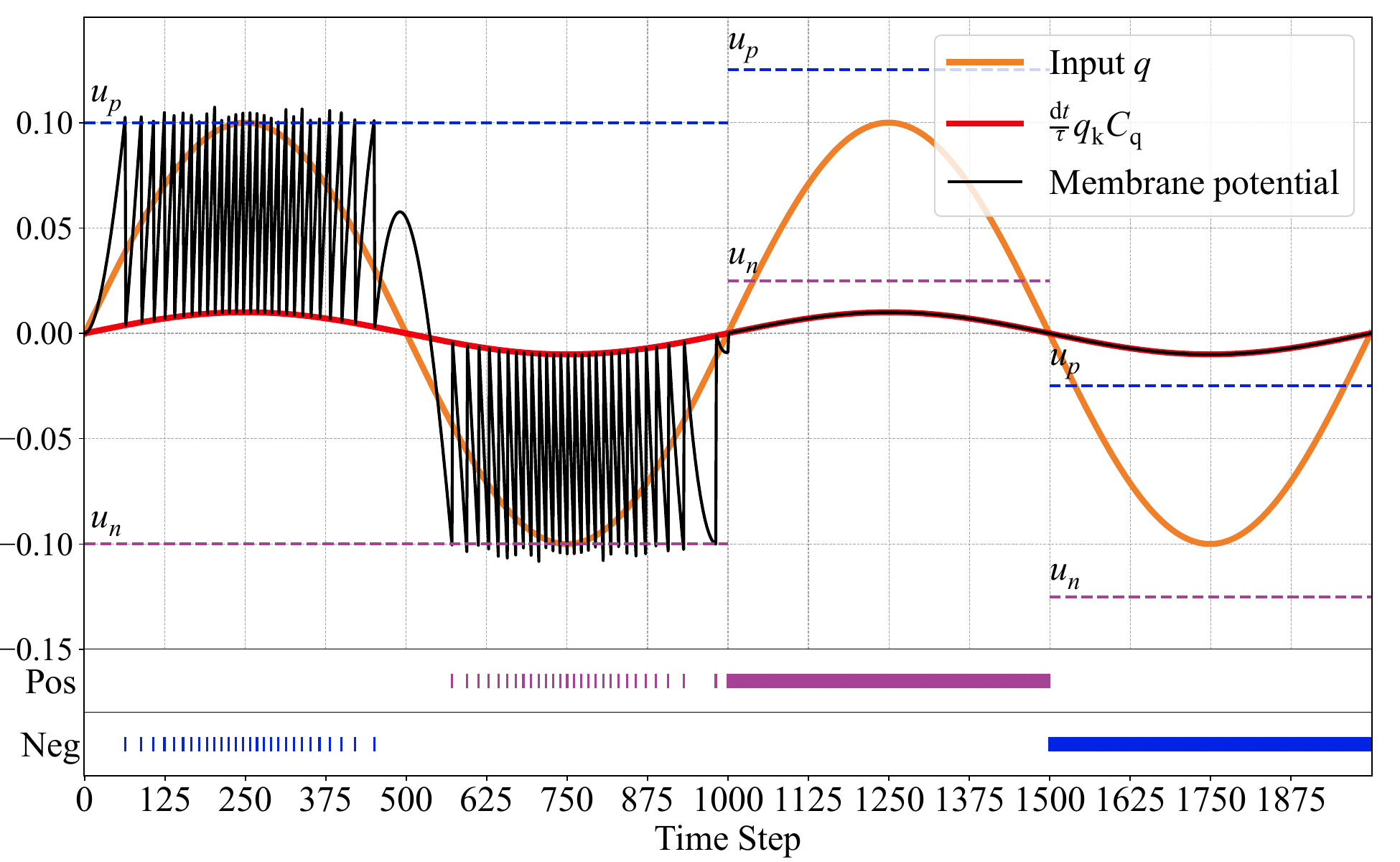}
    \caption{Example of DTS neuron dynamics.
    In this example, the hyperparameter \(C_{\mathrm{q}}\) is 10 and \(\tau\) is 0.1. The sampling interval \(\mathrm{d}t\) is 0.001s and the simulation duration is 2s. The input \(q\) is a sine function shown in orange. The weighted input \(\frac{\mathrm{d}t}{\tau} C_{\mathrm{q}} q^\mathrm{k+1}\) is described in red. The membrane potential \(u\) is black. Both positive and negative spikes are shown as scatter lines.
    This example covers three basic situations of a DTS neuron.}
    \label{fig:sn-illustration}
\end{figure}

Fig.~\ref{fig:sn-illustration} illustrates the dynamics of a DTS neuron, showing three basic behaviors of DTS neurons. 
From 0 to 1 second, $u_{\mathrm{p}}$ is 0.1 and $u_{\mathrm{n}}$ is -0.1. The input gradually accumulates at this stage, reaches any threshold, and falls back. The membrane potential shows discontinuity, and the output spikes are discrete. It should also be noted that the membrane potential and output spikes show symmetry during this period. From 1 to 1.5s, $u_{\mathrm{p}}$ is 0.125 and $u_{\mathrm{n}}$ is 0.025. Because $|o^{\mathrm{k}}|$ is always 1, the neuron dynamics is $u^{\mathrm{k}} = \frac{\mathrm{d}t}{\tau} C_{\mathrm{q}} q^\mathrm{k}$. In this situation, the membrane potential is equal to the input and the DTS neuron continues to generate positive spikes until the input exceeds $u_{\mathrm{n}}$. For 1.5 to 2 seconds, the neuron's behavior is symmetric to the second situation. This example demonstrates that, despite receiving the same inputs, the behavior of the DTS neuron varies significantly under different threshold combinations. This variability enables the control of a soft snake robot. The following section will introduce how DTS is combined with the snake segment.

\subsection{SpikingSoft controller}
\label{methods_contoller}

The symmetry, stability, and diverse output patterns of DTS neurons suggest that the spiking neurons no longer require a pre-designed gait or complex network structure but can control the soft snake segments alone. The temporal characteristics of spiking neurons make this control a closed-loop way. 
As a preliminary introduction, we present the adaptability of spiking neurons to oscillatory nonlinear systems in Fig.~\ref{fig:car_system}. In this figure, a car $m$ on the ground is connected to the origin via a spring $k$ and a damper $d$. The distance $q$ between the car and the origin is the input to a DTS neuron. The neuron's output acts on the car as the force $\tau$. We show the phase portraits in Fig.~\ref{fig:car_system}. 
Different portraits represent the different basic situations of the spiking neuron. Each trajectory depicts a different initial condition of the system. 
The blue symbol means the initial state. Under different initial states, the system will converge to the same limit circle without diverging or collapsing, implying that the system will reach a steady oscillatory state. This phenomenon shows the DTS neuron model can excite and utilize the inner oscillation to form different stable patterns. We generalize this case to soft snake robots, where such oscillations are important to determine the bio-inspired gait. Fig.~\ref{fig:overall} shows the details of the SpikingSoft controller.

\begin{figure*}[t]
    \centering
    \begin{subfigure}[b]{0.76\textwidth}
        \centering
        \includegraphics[width=\linewidth]{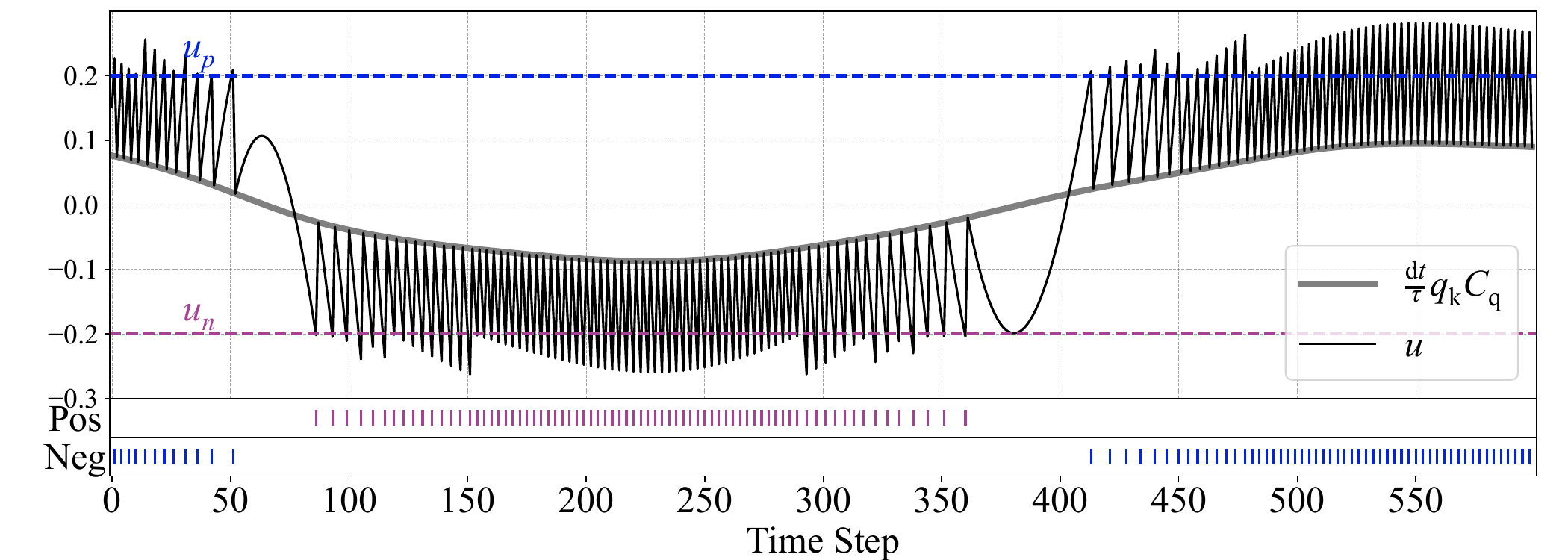}
    \end{subfigure}%
    \hfill
    \begin{subfigure}[b]{0.23\textwidth}
        \centering
        \includegraphics[width=\linewidth]{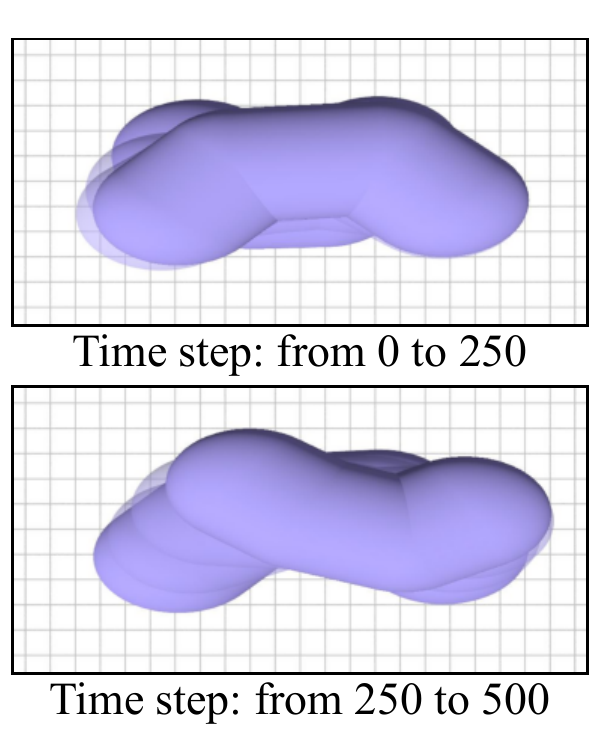}
    \end{subfigure}
    \caption{Example of the DTS neuron (left) and snake segment states with the SpikingSoft controller (right). $u_{\mathrm{p}}$ is 0.2 and $u_{\mathrm{n}}$ is -0.2. At time step 0, the snake segment has an initial positive deformation.}
    \label{fig:snake_case0}
\end{figure*}

\begin{figure*}[t]
    \centering
    \begin{subfigure}[b]{0.54\textwidth}
        \centering
        \includegraphics[width=\linewidth]{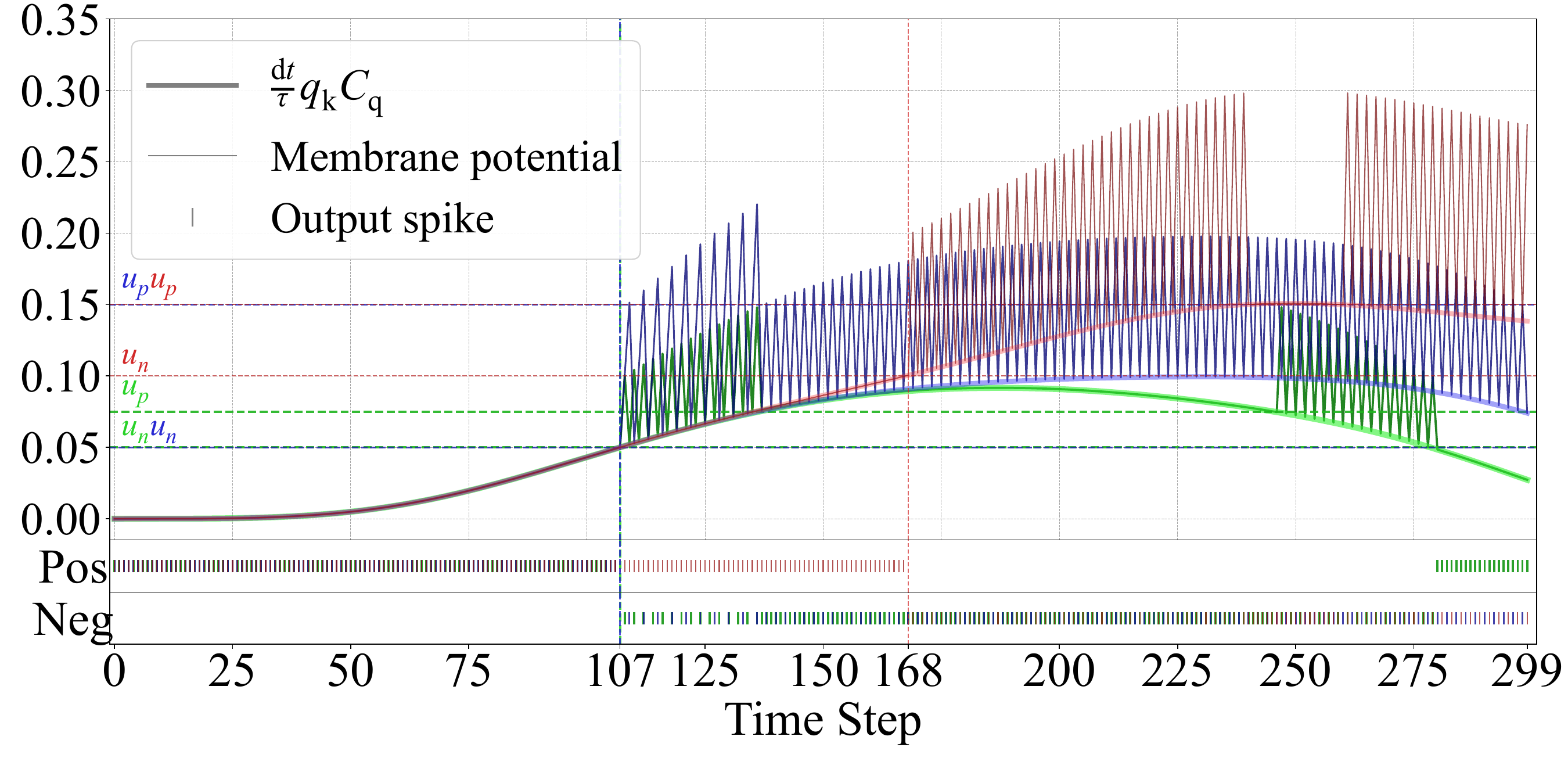}
    \end{subfigure}
    \hfill
    \begin{subfigure}[b]{0.45\textwidth}
        \centering
        \includegraphics[width=\linewidth]{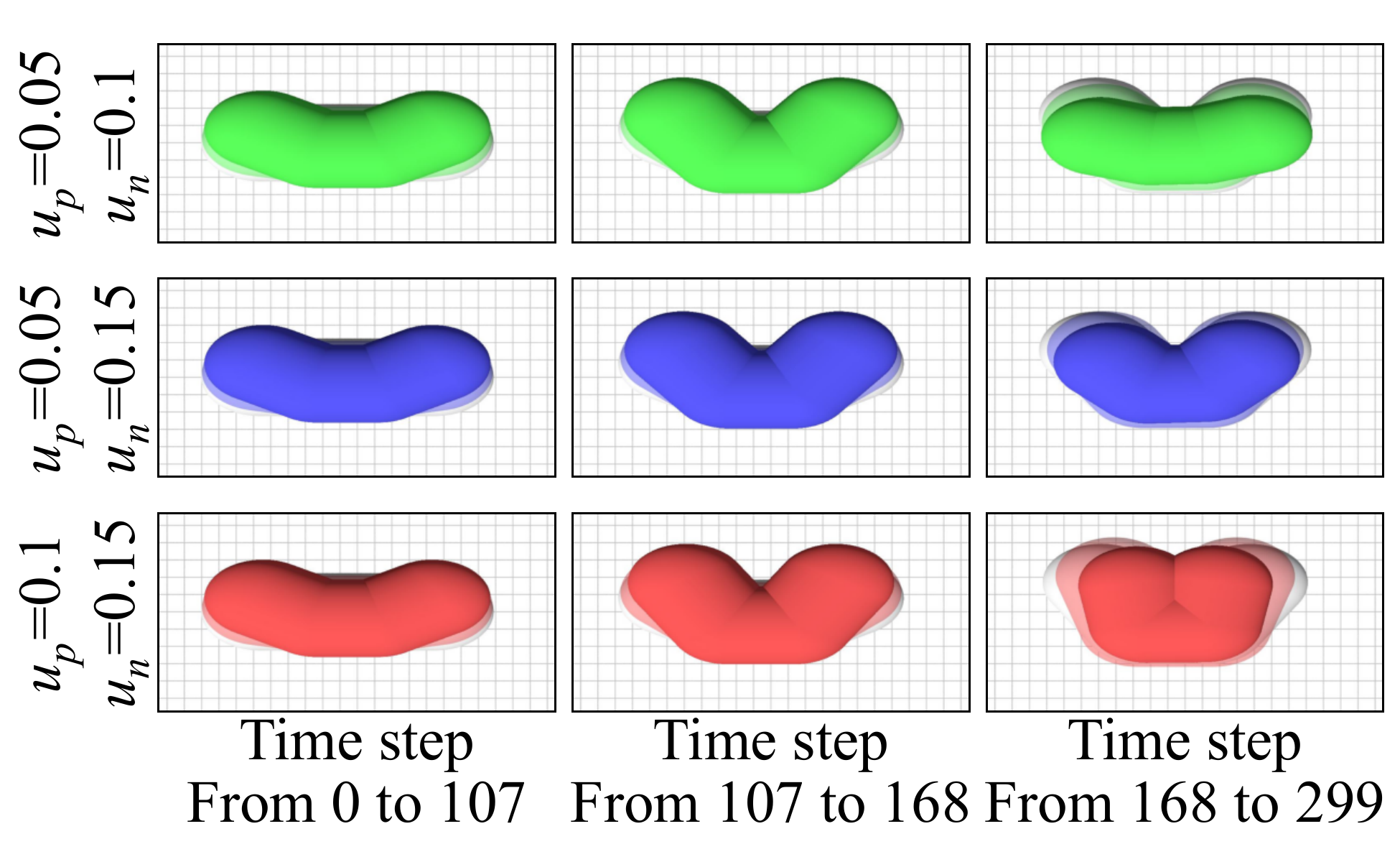}
    \end{subfigure}
    \caption{\textbf{Left}: Example of the DTS neuron state when $u_{\mathrm{n}}$ and $u_{\mathrm{p}}$ are of the same sign. The figure shows three cases: 1) Green represents $u_{\mathrm{n}}$ equals 0.05 and $u_{\mathrm{p}}$ equals 0.075; 2) Blue means $u_{\mathrm{n}}$ equals 0.05 and $u_{\mathrm{p}}$ equals 0.15; 3) Red represents $u_{\mathrm{n}}$ equals 0.1 and $u_{\mathrm{p}}$ equals 0.15. \textbf{Right}: Three examples of the segment state with the SpikingSoft controller when $u_{\mathrm{n}}$ and $u_{\mathrm{p}}$ are of the same sign, which corresponds to left figure.}
    \label{fig:snake_case1}
\end{figure*}

The SpikingSoft controller contains one single DTS neuron, which controls one $n$-node segment of the soft snake. We use Cosserat rod theory and Elastica \cite{zhang2019modeling} to model and simulate the structure of the soft snake in this work. Due to highly flexible materials, the soft snake robot can oscillate during movement, which provides the input $q$ of the SpikingSoft controller. This work uses the deformation $d$ as the oscillating metric. We take the radians between the first node, the $\frac{1}{3}n$ node, and the $\frac{2}{3}n$ node as $d$. The upward deformation sign is positive and the downward deformation sign is negative.
At the time step $k$, the deformation $d^{\mathrm{k}}$ of the segment serves as input $q^{\mathrm{k}}$ of the DTS neuron, and the output $\Gamma^{\mathrm{k}}$ of the DTS neuron works as the input torque of the segment. Since the input torque simulates the biological muscle torque, the segment will also generate a reaction torque $-\Gamma^{\mathrm{k}}$ to achieve force balance. In the SpikingSoft controller, $\Gamma$ acts on the first node of the segment, and $-\Gamma$ acts on the last node.

\begin{algorithm}
    \caption{Pseudocode of the SpikingSoft controller with one snake segment}
    \label{algo.dts}
    \begin{algorithmic}
    \State \textbf{Result: } Updated segment state $s$, updated DTS state $u$, DTS output $o$.
    \State \textbf{Input: } Initial segment state $s$, initial DTS state $u$, initial DTS output $o$, time duration $\mathrm{d}t$, time steps $TimeSteps$, constant $C_{\mathrm{q}}$, $C_{\mathrm{t}}$ and $\tau$, action $a$.
    \State \textbf{Initialize} $u^{\mathrm{k}}=u$, $s^{\mathrm{k}}=s$, $o^{\mathrm{k-1}}=o$
    \State $u_{\mathrm{n}} \leftarrow a[0]-\mathrm{abs}(a[1])$, $u_{\mathrm{p}} \leftarrow a[0]+\mathrm{abs}(a[1])$
    \For{$i \in TimeSteps$}{}
        \State $q^{\mathrm{k}} \leftarrow \mathrm{GetsegmentDeformation}(s^{\mathrm{k}})$
        \State $u^{\mathrm{k}} \leftarrow (1-\mathrm{abs}(o^{\mathrm{k-1}})) (1-(\mathrm{d}t/\tau)) u^{\mathrm{k}} + (\mathrm{d}t/\tau) C_{\mathrm{q}} q^{\mathrm{k}}$
        \If{$u^{\mathrm{k}}>u_\mathrm{p}$}
            \State $o^{\mathrm{k-1}} \leftarrow -1.0$
        \Else
            \State $o^{\mathrm{k-1}} \leftarrow (u^{\mathrm{k}}<u_\mathrm{n})$
        \EndIf
        \State $Torque \leftarrow C_{\mathrm{t}}o^{\mathrm{k-1}}$
        \State $s^{\mathrm{k}} \leftarrow \mathrm{SnakeUpdate}(s^{\mathrm{k}}, Torque, \mathrm{d}t)$
    \EndFor
    \State $s \leftarrow s^{\mathrm{k}}, u \leftarrow u^{\mathrm{k}}, o \leftarrow o^{\mathrm{k-1}}$
    \State \textbf{Return} $s, u, o$
    \end{algorithmic}
\end{algorithm}
The examples in Fig.~\ref{fig:snake_case0} and Fig.~\ref{fig:snake_case1} illustrate how the SpikingSoft controller works with three basic behaviors of DTS neurons. In Fig.~\ref{fig:snake_case0}, $u_{\mathrm{p}}$ is positive and $u_{\mathrm{n}}$ is negative. With this threshold distribution, segment deformation \textit{triggers} spikes. 
When $d$ accumulates to the threshold, the SpikingSoft controller generates torque in the opposite direction to restore the segment deformation. Since the signs of the two thresholds are different, the SpikingSoft controller controls the deformation within a certain range. The segment states in Fig.~\ref{fig:snake_case0} show the segment vibrating in 500 time steps. Meanwhile, the segment moves forward due to environmental friction. 

In Fig.~\ref{fig:snake_case1}, $u_{\mathrm{n}}$ and $u_{\mathrm{p}}$ have the same sign. The segment deformation \textit{stops} spikes with this threshold distribution. The SpikingSoft controller generates spikes until the deformation reaches the threshold at some time. 
The figure shows this process from time step 0 to 168. The segment states are shown in Fig.~\ref{fig:snake_case1}.Right. 
The first column of Fig.~\ref{fig:snake_case1}.Right shows that the segments gradually bend upward from the initial horizontal state as the controller produces positive spikes. At time step 107, the deformations of the green and blue segments reach $u_{\mathrm{n}}$, stopping the positive generation. However, the red segment has a larger $u_{\mathrm{n}}$, which causes it to continue to emit positive spikes until time step 168. This difference is why the deformation of the red segment in the second column is larger than that of the other two segments.

When the deformation reaches $u_{\mathrm{n}}$, the membrane potential gradually accumulates until $u_{\mathrm{p}}$ is reached. The SpikingSoft controller generates the negative spike to recover the segment. The green segment recovers more since the $u_{\mathrm{p}}$ is lower than that of the blue segment. In the third column of Fig.~\ref{fig:snake_case1}.Right, the deformation of the green segment is smaller than that of the blue segment. It should also be noted that although the red segment and the blue segment have the same $u_{\mathrm{p}}$, the red segment produces more positive deformation because its $u_{\mathrm{n}}$ is larger. Therefore, in the last column, the deformation of the red segment is greater than that of the blue segment.

These examples demonstrate the flexibility of the SpikingSoft controller. By simply tuning the thresholds, the controller can generate multiple spike patterns, resulting in various movements in the snake segment. The pseudocode of the SpikingSoft controller is described in Algorithm \ref{algo.dts}. The SpikingSoft controller is also scalable. The $m$-segment snake shown in Fig.~\ref{fig:overall} has $m$ DTS neurons.

\section{Experiments}
\label{experiments}

\begin{figure*}
\vspace{0.2cm}
    \centering
    \begin{minipage}[b]{0.78\linewidth}
    \begin{subfigure}[b]{0.49\textwidth}
        \centering
        \includegraphics[width=\linewidth]{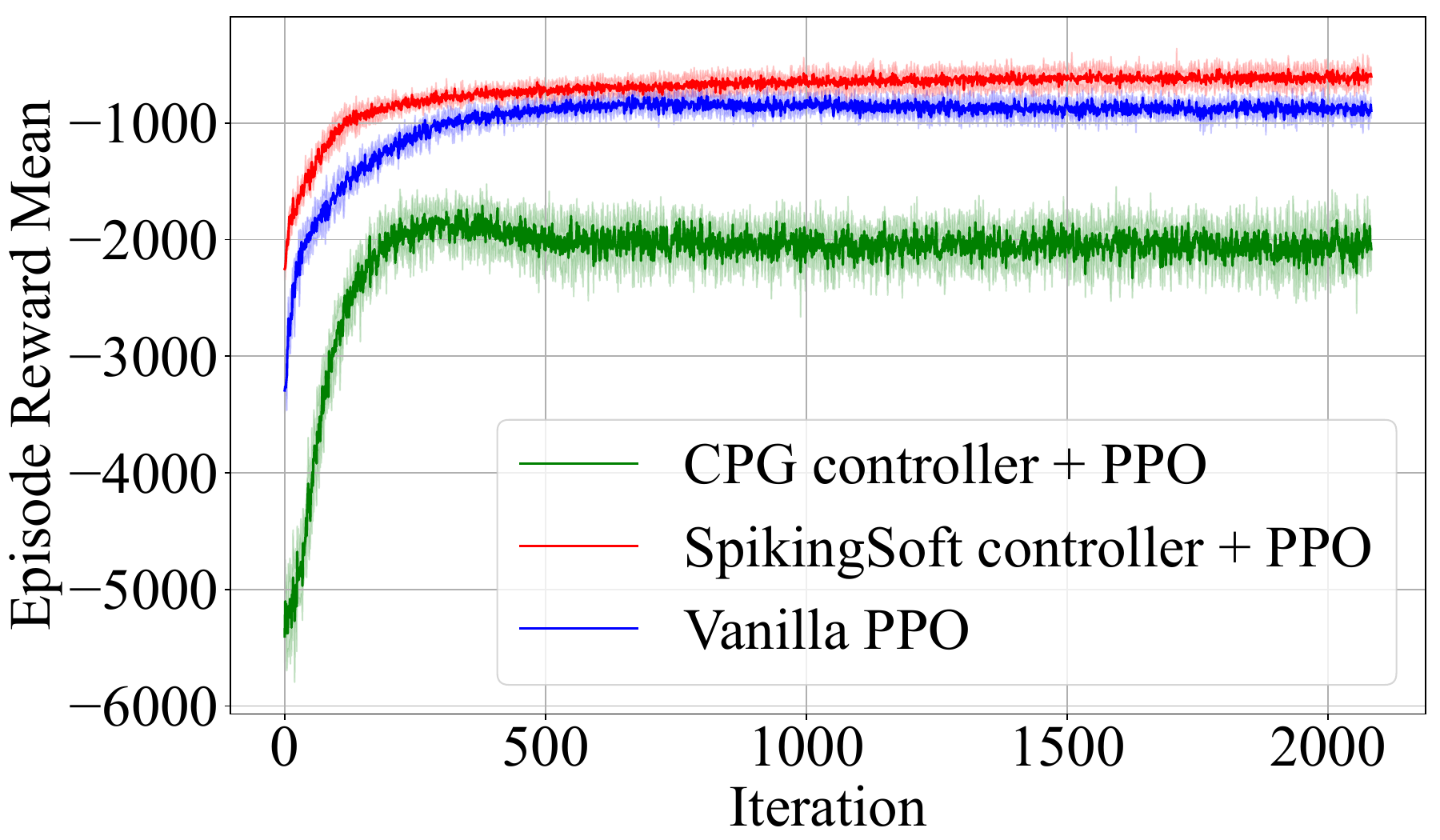}
    \end{subfigure}
    \hfill
    \begin{subfigure}[b]{0.49\textwidth}
        \centering
        \includegraphics[width=\linewidth]{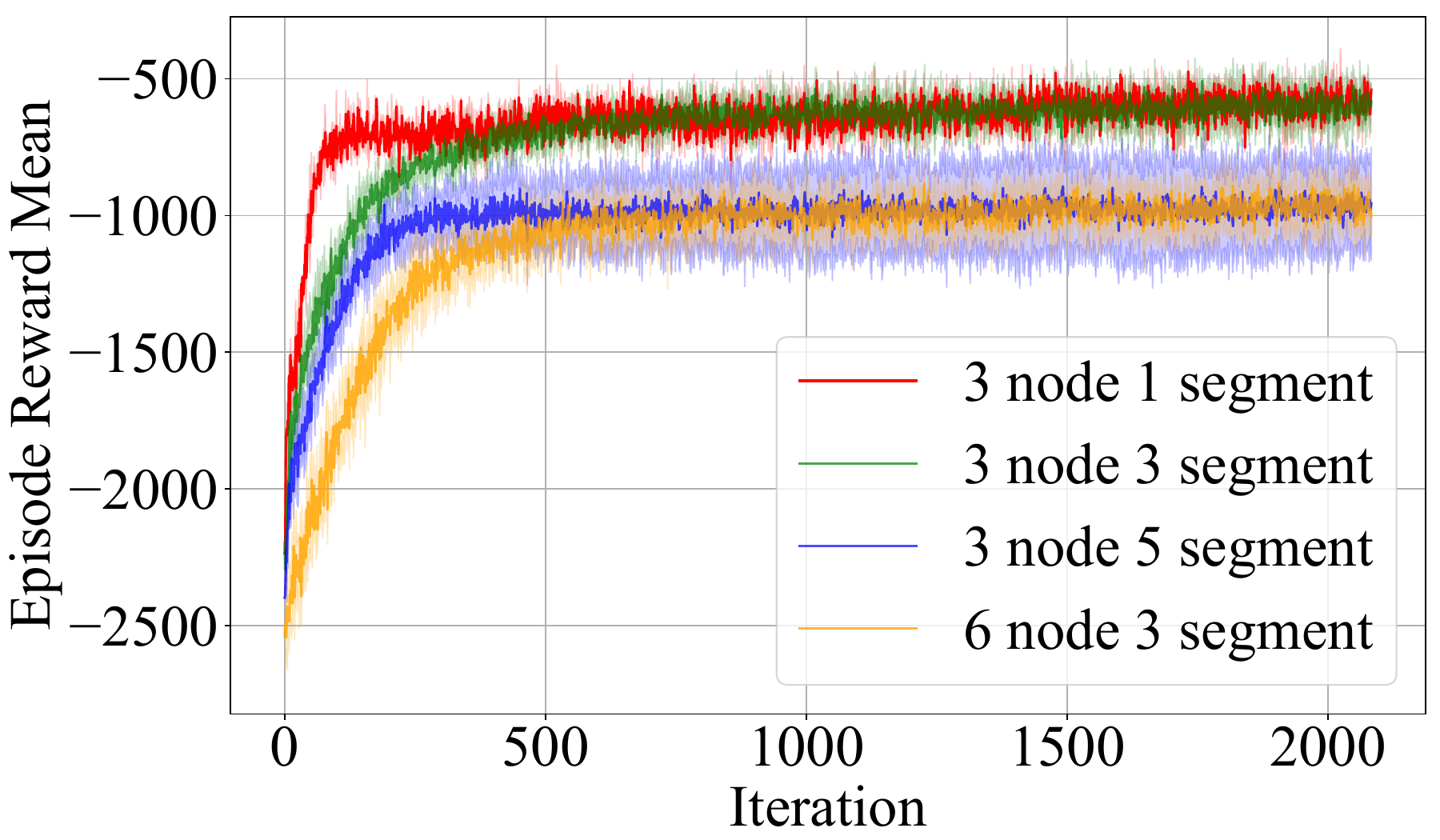}
    \end{subfigure}
    \end{minipage}
    \hfill
    \begin{minipage}[b]{0.21\linewidth}
    \caption{Left: Training reward curve comparison with same PPO framework of three controllers. \\ Right: Training reward curve with different snake settings with the SpikingSoft controller to show the scalability. }
    \label{fig:training_curve}
    \end{minipage}
\vspace{-0.5cm}
\end{figure*}

The experiment aims to control the soft snake robot to reach a target object on a two-dimensional plane. In this work, the snake model takes three-dimensional Cosserat rod Theory, which supports various effects including bending, twisting, shear and elongation. The environment incorporates anisotropic friction during locomotion. We select a 3-segment 3-node snake, which means $n$ and $m$ are 3. The radius of each node is set to 0.25. The starting point of the snake is the origin, and the direction is parallel to the negative $x$ axis. The target position is randomly sampled from a circle with a center of $(4, 0)$ and a radius of 8.
More experiment parameters are shown in Tab.~\ref{table:para}.

\begin{table}[!t]
\begin{tabularx}{\linewidth}{@{} *{2}{X} @{}}
\toprule
Parameter & Value (SI Units) \\
\midrule
$C_{\mathrm{q}}$ & 5.0 \\
$C_{\mathrm{t}}$ & 0.1 \\
$\tau$ & 0.1 \\
\midrule
Density & 1.0 \\
Rayleigh Damping coefficient & 2e-3 \\
Poisson ratio & 0.5 \\
Young's modulus & 50 \\
Gravity & -9.80665 \\
Friction Froude & 0.1 \\
$\mu$ & ${\mathrm{Length}}/{0.1\mathrm{Gravity}}$\\
Anisotropic friction & $[1.0 \mu , 0.0001 \mu, 1.0 \mu]$ \\
\midrule
$\mathrm{d}t$ & 0.001 \\
\bottomrule
\end{tabularx}
\caption{Experiment parameters.}
\label{table:para}
\end{table}

We use RL algorithms to train the SpikingSoft controller as shown in Fig.~\ref{fig:overall}. The RL agent update frequency is 2Hz. The observation state has 23 dimensions, expressed as $Obs = [Obs_{\mathrm{x}}, Obs_{\mathrm{y}}, Obs_{\mathrm{u}}]$, where $Obs_{\mathrm{x}} = [x_{\mathrm{target}} - x_{\mathrm{i}} \text{, for } i \in 0, ..., mn+1]$, $Obs_{\mathrm{y}} = [y_{\mathrm{target}} - y_{\mathrm{i}} \text{, for } i \in 0, ..., mn+1]$, and $Obs_{\mathrm{u}} = [u_{\mathrm{j}} \text{, for } j \in 0,...,m]$.
$(x_{\mathrm{target}}, y_{\mathrm{target}})$ is the 2D position of the target object. $(x_{\mathrm{i}}, y_{\mathrm{i}})$ is the position of the snake's $i^{th}$ node. $u_{\mathrm{j}}$ is the membrane potential of the DTS neuron in the $j^{th}$ segment. The action of the SpikingSoft controller has six dimensions, which are described as $action = [\mu_{\mathrm{i}}, \sigma_{\mathrm{i}} \text{, for } i \in 0,...,m]$, where $\mu$ represents the mean value and $\sigma$ represents the interval value. So $u_{\mathrm{p}}^{\mathrm{i}} = \mu_{\mathrm{i}} + |\sigma_{\mathrm{i}}|$ and $u_{\mathrm{n}}^{\mathrm{i}} = \mu_{\mathrm{i}} - |\sigma_{\mathrm{i}}|$. $i$ means the $i^{th}$ segment of the snake. We set both $\mu$ and $\sigma$ in the range of [-3.25, 3.25]. During training, the reward function is set as $r = r_{\mathrm{1}} + r_{\mathrm{2}} + r_{\mathrm{3}} + r_{\mathrm{4}}$:
\begin{equation}
\begin{aligned}
& r_{\mathrm{1}}  = \left\{ 
\begin{array}{ll}
     1, \text{ if } l < 1, \\
     5, \text{ else if } l < 0.5, \\
     10, \text{ else if } l < 0.25; 
\end{array}
\right. \text{ }
r_{\mathrm{2}}  = 250 \text{ if } l < 0.1; \text{ } \\
& r_{\mathrm{3}}  = - l^2; \text{ }
r_{\mathrm{4}} = -5000 \text{ if destruction,} 
\end{aligned}
\end{equation}
where $l = \sqrt{{(x_{\mathrm{target}} - x_{\mathrm{0}})}^2 + {(y_{\mathrm{target}} - y_{\mathrm{0}})}^2}$. 

$r_{\mathrm{1}}$ is the range reward that incentivizes the snake to move closer to the target. Within a range of distances of $[0, 1]$, the reward increases as the snake's head approaches the target. $r_{\mathrm{2}}$ is the goal-reaching reward, which provides significant positive reinforcement when the snake reaches the target. $r_{\mathrm{3}}$ is the step penalty which discourages unnecessary movements by imposing a small penalty as each time step. $r_{\mathbf{4}}$ is the destruction penalty, which applies a large negative reward when the snake encounters an invalid state.

\begin{table*}
\vspace{0.2cm}
\begin{tabularx}{\linewidth}{@{} >{\hsize=0.3\hsize}X >{\hsize=0.175\hsize}X >{\hsize=0.175\hsize}X >{\hsize=0.175\hsize}X >{\hsize=0.175\hsize}X @{}}
\toprule
Algorithm & Success Rate $\uparrow$ & Game Time (s) $\downarrow$ & Total Reward $\uparrow$ & Silence Rate $\uparrow$ \\ 
\midrule
\midrule
Random action  & 9.70\% ± 2.28\% & 46.71 ± 10.19 & -13769.06 ± 9274.31 & 0.0\% \\
\midrule
Vanilla with PPO   & 66.00\% ± 6.71\%   & 26.95 ± 17.60 & -2391.28 ± 3745.52  & 0.0\% \\ 
CPG controller with PPO & 9.30\% ± 3.13\% & 47.03 ± 9.90 & -4462.13 ± 3728.16 & 13.46\% ± 6.57\% \\
SpikingSoft controller with PPO (\textbf{Ours}) & \textbf{87.60\% ± 3.32\%}   & \textbf{18.96 ± 12.15} & \textbf{-843.81 ± 799.27}  & \textbf{28.48\% ± 8.63\%} \\
\bottomrule
\end{tabularx}
\caption{Performance comparison of different methods.}
\label{table:result}
\end{table*}
\begin{table*}[]
\begin{tabularx}{\linewidth}{@{} >{\hsize=0.3\hsize}X >{\hsize=0.175\hsize}X >{\hsize=0.175\hsize}X >{\hsize=0.175\hsize}X >{\hsize=0.175\hsize}X @{}}
\toprule
    Snake Size & Success Rate & Game Time & Total Reward & Silence Rate \\
\midrule
\midrule
    3 node 1 segment & 84.00\% ± 2.16\% & 22.09 ± 13.05 & -920.43 ± 1072.01 & 7.15\% ± 4.06\%\\
    3 node 3 segment & 89.33\% ± 0.47\% & 19.24 ± 11.09  & -805.18 ± 669.44 & 26.42\% + 7.57\%  \\
    3 node 5 segment & 79.33\% ± 3.40\% & 23.01 ± 14.34 & -1043.82 ± 976.00 & 49.59\% ± 7.50\% \\
    6 node 3 segment & 91.33\% ± 0.94\% & 21.34 ± 9.88 & -1004.14 ± 770.97 & 21.46\% ± 5.14\%\\
\bottomrule
\end{tabularx}
\caption{Performance comparison of different settings of the snake.}
\label{table:scalability}
\end{table*}

The comparison of the training average reward curve is in Fig.~\ref{fig:training_curve}.Left. The vanilla and SpikingSoft methods show similar trends, where the episode's average reward increases initially and stabilizes during iterations. Compared to the vanilla method, the SpikingSoft controller significantly enhances learning and performance. However, the CPG controller does not work well in this scenario. This may be because the oscillation generated by the CPG cannot translate into effective and stable movement of the snake body due to its rich dynamics. 
We use the PPO algorithm \cite{schulman2017proximal} with MLP policies to train the agent and compare the cases of using the SpikingSoft controller, the CPG controller\cite{ijspeert2008central}, and not using the controllers (i.e., vanilla RL), respectively. The MLP policy has two layers and each contains 64 neurons. In the CPG controller, the same structure in \cite{liu2023reinforcement} is used,
and the actions are the tonic inputs of the CPG oscillators. To determine the hyperparameters in the CPG network, Genetic Algorithm is employed with the same experiment in \cite{liu2023reinforcement}. The amplitude ratio is 2.30. The self and mutual inhibit weights are 10.05 and 2.18, respectively. The discharge rate is 0.56 and the adaption rate is 1.76. The coupling weights are 9.13 and 0.73, respectively. With the vanilla RL controller, the actions are $action = [\Gamma_{\mathrm{i}} \text{ for } i \in 0,...,m]$ where $\Gamma$ has a range of [-50, 50] and the dimension of the observation space of the environment becomes $Obs = [Obs_{\mathrm{x}}, Obs_{\mathrm{y}}]$. To align with the SpikingSoft controller, in all baseline methods, the output torque still acts on the first node of each segment, and a reaction torque is also generated on the last node. Other details of RL training remain the same for comparison. The training process contains $10^7$ time steps. We trained five times for each method.
To demonstrate the scalability of the proposed SpikingSoft controller, we introduce additional scenarios with varying snake lengths. Fig.~\ref{fig:training_curve}.Right presents the training curves for four distinct configurations: a short snake with 3 nodes and 1 segment, the standard 3-node 3-segment snake, an extended version with 3 nodes and 5 segments, and a larger snake with 6 nodes and 3 segments. The results show that all training processes successfully converge, highlighting the robustness of the controller across different snake lengths.

\begin{figure*}[]
    \centering
    \includegraphics[width=\textwidth]{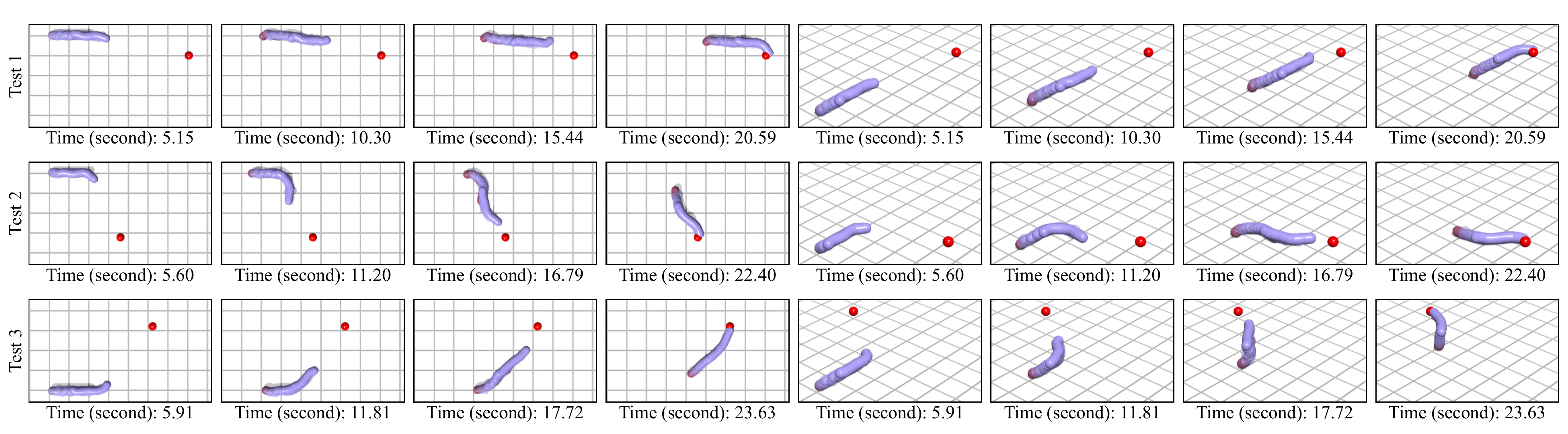}
    \caption{Example of evaluation scenarios for the 3-segment 3-node snake with the SpikingSoft controller. The purple snake moves toward the red sphere on a 2D plane.}
    \label{fig:result_visualization}
\end{figure*}
We also evaluate the methods on the test scenarios, where the goal of the task is to touch the target object with a radius of 0.25 within 50 seconds of game time. Performance comparison is illustrated in Tab.~\ref{table:result}. Comparison metrics include success rate, average game time, and average total episode reward. We tested 1000 episodes with randomly sampled target positions for each method. As shown in the table, the test results also fit the training curves. Note that since the game time during testing is more extended than during training, the average test reward is smaller than that of training. The result suggests that the SpikingSoft controller offers a higher success rate, efficient game completion, and higher rewards. The test results of four different configurations of snakes are in Tab.~\ref{table:scalability}, where all snakes achieve a high success rate.
Additionally, we introduce a metric, the Silence Rate, to evaluate the activity levels of the controllers. In the SpikingSoft controller, the DTS neurons exhibit event-triggered behavior, meaning they do not generate output torques at every time step. To quantify this, we calculate the rate at which the controller produces zero output as the Silence Rate. A higher Silence Rate is desirable as it suggests greater energy efficiency. As shown in Tab.~\ref{table:result}, the SpikingSoft controller achieves a significantly higher Silence Rate compared to the CPG controller, highlighting the advantages of the event-based mechanism in reducing energy consumption.

\begin{figure*}
    \centering
    \includegraphics[width=\linewidth]{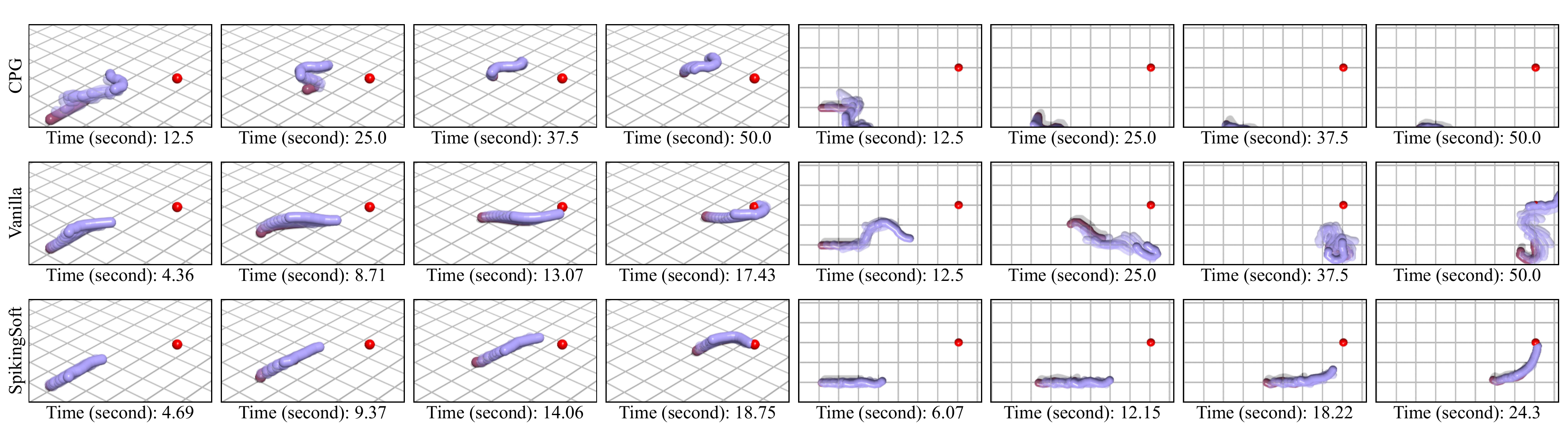}
    \caption{Test scenario performance comparison of the CPG, vanilla RL, and SpikingSoft controllers with MLP PPO. The first to fourth columns are top views, and the fifth to eighth are diagonal views.}
    \label{fig:result_visualization_compare}
\end{figure*}

Fig.~\ref{fig:result_visualization} visualizes the movement of the 3-segment 3-node snake with the SpikingSoft controller. The purple snake can change direction smoothly and touch target objects at different positions.
We show the visualization comparison between the SpikingSoft, CPG, and vanilla RL controllers in Fig.~\ref{fig:result_visualization_compare}. 
The four rows on the left compare the scenario where the target position is $(9, -4)$. Both the vanilla RL controller and the SpikingSoft controller successfully make the snake reach the target. However, it is observed that the snake controlled by SpikingSoft tends to move forward with a smooth wiggle. In contrast, the snake controlled by the vanilla controller has more after-images, meaning it fluctuates more as it moves forward. The four rows on the right compare the scenarios where the target position is $(12.5, 5)$. In this scenario, only the SpikingSoft controller can still reach the target object well in 24.3 seconds. However, the other two controllers took 50 seconds and failed. 

\section{Conclusion}
\label{conclusion}
We propose a DTS neuron and a SpikingSoft controller for the soft snake robot in the locomotion task. Simulation experiments have shown that the SpikingSoft controller can allow the snake to reach the target object faster and smoother. Compared to the vanilla method, the DTS neuron plays an important role in stabilizing control. This is because the design of double thresholds gives it the chance of rapid recovery from huge deformations. Although this work used deformation as input to DTS neurons, other oscillatory factors during movement can also be used, providing great possibilities for real-world applications in future work. Meanwhile, the DTS neuron can emit positive and negative spikes alternately, which allows the snake to move zigzag without a known serpenoid curve. This work paves the way for substantial future research into bio-inspired controllers for bionic soft robots.

\textbf{Limitations} 
Although the SpikingSoft exhibits commendable efficacy and success in simulation environments, its application to actual robotic experiments remains to be further explored. Future research will focus on implementing the SpikingSoft controller in a practical setting, specifically for locomotion tasks in a real soft snake robot.

\bibliographystyle{IEEEtran}
\bibliography{refs.bib}

\begin{thebibliography}{10}
\providecommand{\url}[1]{#1}
\csname url@samestyle\endcsname
\providecommand{\newblock}{\relax}
\providecommand{\bibinfo}[2]{#2}
\providecommand{\BIBentrySTDinterwordspacing}{\spaceskip=0pt\relax}
\providecommand{\BIBentryALTinterwordstretchfactor}{4}
\providecommand{\BIBentryALTinterwordspacing}{\spaceskip=\fontdimen2\font plus
\BIBentryALTinterwordstretchfactor\fontdimen3\font minus \fontdimen4\font\relax}
\providecommand{\BIBforeignlanguage}[2]{{%
\expandafter\ifx\csname l@#1\endcsname\relax
\typeout{** WARNING: IEEEtran.bst: No hyphenation pattern has been}%
\typeout{** loaded for the language `#1'. Using the pattern for}%
\typeout{** the default language instead.}%
\else
\language=\csname l@#1\endcsname
\fi
#2}}
\providecommand{\BIBdecl}{\relax}
\BIBdecl

\bibitem{kuo2023deep}
P.-H. Kuo, C.-H. Pao, E.-Y. Chang, and H.-T. Yau, ``Deep-reinforcement-learning-based gait pattern controller on an uneven terrain for humanoid robots,'' \emph{International Journal of Optomechatronics}, vol.~17, no.~1, p. 2222146, 2023.

\bibitem{khan2023review}
M.~S. Khan and R.~K. Mandava, ``A review on gait generation of the biped robot on various terrains,'' \emph{Robotica}, vol.~41, no.~6, pp. 1888--1930, 2023.

\bibitem{qi2020novel}
X.~Qi, H.~Shi, T.~Pinto, and X.~Tan, ``A novel pneumatic soft snake robot using traveling-wave locomotion in constrained environments,'' \emph{IEEE Robotics and Automation Letters}, vol.~5, no.~2, pp. 1610--1617, 2020.

\bibitem{seetohul2022snake}
J.~Seetohul and M.~Shafiee, ``Snake robots for surgical applications: A review,'' \emph{Robotics}, vol.~11, no.~3, p.~57, 2022.

\bibitem{liu2021review}
J.~Liu, Y.~Tong, and J.~Liu, ``Review of snake robots in constrained environments,'' \emph{Robotics and Autonomous Systems}, vol. 141, p. 103785, 2021.

\bibitem{della2020soft}
C.~Della~Santina, M.~G. Catalano, A.~Bicchi, M.~Ang, O.~Khatib, and B.~Siciliano, ``Soft robots,'' \emph{Encyclopedia of Robotics}, vol. 489, 2020.

\bibitem{rajashekhar2024developments}
V.~Rajashekhar, G.~Prabhakar \emph{et~al.}, ``Developments and trend maps for soft continuum manipulators and soft snake robots,'' \emph{Foundations and Trends{\textregistered} in Robotics}, vol.~12, no.~1, pp. 1--74, 2024.

\bibitem{ma1999analysis}
S.~Ma, ``Analysis of snake movement forms for realization of snake-like robots,'' in \emph{Proceedings 1999 IEEE International Conference on Robotics and Automation (Cat. No. 99CH36288C)}, vol.~4.\hskip 1em plus 0.5em minus 0.4em\relax Ieee, 1999, pp. 3007--3013.

\bibitem{li2021adaptive}
D.~Li, Z.~Pan, H.~Deng, and L.~Hu, ``Adaptive path following controller of a multijoint snake robot based on the improved serpenoid curve,'' \emph{IEEE Transactions on Industrial Electronics}, vol.~69, no.~4, pp. 3831--3842, 2021.

\bibitem{arachchige2021soft}
D.~D. Arachchige, Y.~Chen, and I.~S. Godage, ``Soft robotic snake locomotion: Modeling and experimental assessment,'' in \emph{2021 IEEE 17th International Conference on Automation Science and Engineering (CASE)}.\hskip 1em plus 0.5em minus 0.4em\relax IEEE, 2021, pp. 805--810.

\bibitem{zhao2021multigait}
W.~Zhao, J.~Wang, and Y.~Fei, ``A multigait continuous flexible snake robot for locomotion in complex terrain,'' \emph{IEEE/ASME Transactions on Mechatronics}, vol.~27, no.~5, pp. 3751--3761, 2021.

\bibitem{ijspeert2008central}
A.~J. Ijspeert, ``Central pattern generators for locomotion control in animals and robots: a review,'' \emph{Neural networks}, vol.~21, no.~4, pp. 642--653, 2008.

\bibitem{wu2010cpg}
X.~Wu and S.~Ma, ``Cpg-based control of serpentine locomotion of a snake-like robot,'' \emph{Mechatronics}, vol.~20, no.~2, pp. 326--334, 2010.

\bibitem{jia2021coach}
Y.~Jia and S.~Ma, ``A coach-based bayesian reinforcement learning method for snake robot control,'' \emph{IEEE Robotics and Automation Letters}, vol.~6, no.~2, pp. 2319--2326, 2021.

\bibitem{liu2023energy}
Y.~Liu and A.~Barati~Farimani, ``An energy-saving snake locomotion pattern learned in a physically constrained environment with online model-based policy gradient method,'' \emph{Journal of Mechanisms and Robotics}, vol.~15, no.~4, p. 041007, 2023.

\bibitem{bing2022simulation}
Z.~Bing, L.~Cheng, K.~Huang, and A.~Knoll, ``Simulation to real: learning energy-efficient slithering gaits for a snake-like robot,'' \emph{IEEE Robotics \& Automation Magazine}, vol.~29, no.~4, pp. 92--103, 2022.

\bibitem{shi2020deep}
J.~Shi, T.~Dear, and S.~D. Kelly, ``Deep reinforcement learning for snake robot locomotion,'' \emph{IFAC-PapersOnLine}, vol.~53, no.~2, pp. 9688--9695, 2020.

\bibitem{qiu2021reinforcement}
K.~Qiu, H.~Zhang, Y.~Lv, Y.~Wang, C.~Zhou, and R.~Xiong, ``Reinforcement learning of serpentine locomotion for a snake robot,'' in \emph{2021 IEEE International Conference on Real-time Computing and Robotics (RCAR)}.\hskip 1em plus 0.5em minus 0.4em\relax IEEE, 2021, pp. 468--473.

\bibitem{liu2023reinforcement}
X.~Liu, C.~D. Onal, and J.~Fu, ``Reinforcement learning of cpg-regulated locomotion controller for a soft snake robot,'' \emph{IEEE Transactions on Robotics}, 2023.

\bibitem{bellegarda2022cpg}
G.~Bellegarda and A.~Ijspeert, ``Cpg-rl: Learning central pattern generators for quadruped locomotion,'' \emph{IEEE Robotics and Automation Letters}, vol.~7, no.~4, pp. 12\,547--12\,554, 2022.

\bibitem{shafiee2024manyquadrupeds}
M.~Shafiee, G.~Bellegarda, and A.~Ijspeert, ``Manyquadrupeds: Learning a single locomotion policy for diverse quadruped robots,'' in \emph{2024 IEEE International Conference on Robotics and Automation (ICRA)}.\hskip 1em plus 0.5em minus 0.4em\relax IEEE, 2024, pp. 3471--3477.

\bibitem{bello2024motor}
S.~Bello-Rojas and M.~W. Bagnall, ``Motor control: Snake neurons speed up,'' \emph{Current Biology}, vol.~34, no.~3, pp. R98--R99, 2024.

\bibitem{schuman2022opportunities}
C.~D. Schuman, S.~R. Kulkarni, M.~Parsa, J.~P. Mitchell, B.~Kay \emph{et~al.}, ``Opportunities for neuromorphic computing algorithms and applications,'' \emph{Nature Computational Science}, vol.~2, no.~1, pp. 10--19, 2022.

\bibitem{abadia2021cerebellar}
I.~Abad{\'\i}a, F.~Naveros, E.~Ros, R.~R. Carrillo, and N.~R. Luque, ``A cerebellar-based solution to the nondeterministic time delay problem in robotic control,'' \emph{Science Robotics}, vol.~6, no.~58, p. eabf2756, 2021.

\bibitem{brockman2016openai}
G.~Brockman, V.~Cheung, L.~Pettersson, J.~Schneider, J.~Schulman, J.~Tang, and W.~Zaremba, ``Openai gym,'' \emph{arXiv preprint arXiv:1606.01540}, 2016.

\bibitem{chen2022deep}
D.~Chen, P.~Peng, T.~Huang, and Y.~Tian, ``Deep reinforcement learning with spiking q-learning,'' \emph{arXiv preprint arXiv:2201.09754}, 2022.

\bibitem{akl2021porting}
M.~Akl, Y.~Sandamirskaya, F.~Walter, and A.~Knoll, ``Porting deep spiking q-networks to neuromorphic chip loihi,'' in \emph{International Conference on Neuromorphic Systems 2021}, 2021, pp. 1--7.

\bibitem{tang2021deep}
G.~Tang, N.~Kumar, R.~Yoo, and K.~Michmizos, ``Deep reinforcement learning with population-coded spiking neural network for continuous control,'' in \emph{Conference on Robot Learning}.\hskip 1em plus 0.5em minus 0.4em\relax PMLR, 2021, pp. 2016--2029.

\bibitem{chen2024fully}
D.~Chen, P.~Peng, T.~Huang, and Y.~Tian, ``Fully spiking actor network with intralayer connections for reinforcement learning,'' \emph{IEEE Transactions on Neural Networks and Learning Systems}, 2024.

\bibitem{bing2018end}
Z.~Bing, C.~Meschede, K.~Huang, G.~Chen, F.~Rohrbein, M.~Akl, and A.~Knoll, ``End to end learning of spiking neural network based on r-stdp for a lane keeping vehicle,'' in \emph{2018 IEEE international conference on robotics and automation (ICRA)}.\hskip 1em plus 0.5em minus 0.4em\relax IEEE, 2018, pp. 4725--4732.

\bibitem{liu2023spiking}
Y.~Liu and W.~Pan, ``Spiking neural-networks-based data-driven control,'' \emph{Electronics}, vol.~12, no.~2, p. 310, 2023.

\bibitem{jiang2020target}
Z.~Jiang, R.~Otto, Z.~Bing, K.~Huang, and A.~Knoll, ``Target tracking control of a wheel-less snake robot based on a supervised multi-layered snn,'' in \emph{2020 IEEE/RSJ International Conference on Intelligent Robots and Systems (IROS)}.\hskip 1em plus 0.5em minus 0.4em\relax IEEE, 2020, pp. 7124--7130.

\bibitem{ding2022biologically}
J.~Ding, B.~Dong, F.~Heide, Y.~Ding, Y.~Zhou, B.~Yin, and X.~Yang, ``Biologically inspired dynamic thresholds for spiking neural networks,'' \emph{Advances in Neural Information Processing Systems}, vol.~35, pp. 6090--6103, 2022.

\bibitem{zhang2023low}
A.~Zhang, J.~Shi, J.~Wu, Y.~Zhou, and W.~Yu, ``Low latency and sparse computing spiking neural networks with self-driven adaptive threshold plasticity,'' \emph{IEEE Transactions on Neural Networks and Learning Systems}, 2023.

\bibitem{yu2021constructing}
Q.~Yu, C.~Ma, S.~Song, G.~Zhang, J.~Dang, and K.~C. Tan, ``Constructing accurate and efficient deep spiking neural networks with double-threshold and augmented schemes,'' \emph{IEEE Transactions on Neural Networks and Learning Systems}, vol.~33, no.~4, pp. 1714--1726, 2021.

\bibitem{rostro2015cpg}
H.~Rostro-Gonzalez, P.~A. Cerna-Garcia, G.~Trejo-Caballero, C.~H. Garcia-Capulin, M.~A. Ibarra-Manzano, J.~G. Avina-Cervantes, and C.~Torres-Huitzil, ``A cpg system based on spiking neurons for hexapod robot locomotion,'' \emph{Neurocomputing}, vol. 170, pp. 47--54, 2015.

\bibitem{tieck2019combining}
J.~C.~V. Tieck, J.~Rutschke, J.~Kaiser, M.~Schulze, T.~Buettner, D.~Reichard, A.~Roennau, and R.~Dillmann, ``Combining spiking motor primitives with a behaviour-based architecture to model locomotion for six-legged robots,'' in \emph{2019 IEEE/RSJ International Conference on Intelligent Robots and Systems (IROS)}.\hskip 1em plus 0.5em minus 0.4em\relax IEEE, 2019, pp. 4161--4168.

\bibitem{wang2019locomotion}
M.~Wang, X.~Li, Y.~Zhang, Z.~Chang, and J.~Yu, ``Locomotion control of robotic fish with a hierarchical framework combining spiking neural networks and cpgs,'' in \emph{2019 IEEE 9th Annual International Conference on CYBER Technology in Automation, Control, and Intelligent Systems (CYBER)}.\hskip 1em plus 0.5em minus 0.4em\relax IEEE, 2019, pp. 1187--1190.

\bibitem{lele2020learning}
A.~S. Lele, Y.~Fang, J.~Ting, and A.~Raychowdhury, ``Learning to walk: Bio-mimetic hexapod locomotion via reinforcement-based spiking central pattern generation,'' \emph{IEEE Journal on Emerging and Selected Topics in Circuits and Systems}, vol.~10, no.~4, pp. 536--545, 2020.

\bibitem{lele2021end}
A.~Lele, Y.~Fang, J.~Ting, and A.~Raychowdhury, ``An end-to-end spiking neural network platform for edge robotics: From event-cameras to central pattern generation,'' \emph{IEEE Transactions on Cognitive and Developmental Systems}, vol.~14, no.~3, pp. 1092--1103, 2021.

\bibitem{yamazaki2022spiking}
K.~Yamazaki, V.-K. Vo-Ho, D.~Bulsara, and N.~Le, ``Spiking neural networks and their applications: A review,'' \emph{Brain Sciences}, vol.~12, no.~7, p. 863, 2022.

\bibitem{jiang2023fully}
X.~Jiang, Q.~Zhang, J.~Sun, and R.~Xu, ``Fully spiking neural network for legged robots,'' \emph{arXiv preprint arXiv:2310.05022}, 2023.

\bibitem{chen2017toward}
G.~Chen, Z.~Bing, F.~R{\"o}hrbein, J.~Conradt, K.~Huang, L.~Cheng, Z.~Jiang, and A.~Knoll, ``Toward brain-inspired learning with the neuromorphic snake-like robot and the neurorobotic platform,'' \emph{IEEE Transactions on Cognitive and Developmental Systems}, vol.~11, no.~1, pp. 1--12, 2017.

\bibitem{wu2018spatio}
Y.~Wu, L.~Deng, G.~Li, and L.~Shi, ``Spatio-temporal backpropagation for training high-performance spiking neural networks,'' \emph{Frontiers in neuroscience}, vol.~12, p. 323875, 2018.

\bibitem{zhang2019modeling}
\BIBentryALTinterwordspacing
X.~Zhang, F.~Chan, T.~Parthasarathy, and M.~Gazzola, ``Modeling and simulation of complex dynamic musculoskeletal architectures,'' \emph{Nature Communications}, vol.~10, no.~1, pp. 1--12, 2019. [Online]. Available: \url{https://doi.org/10.1038/s41467-019-12759-5}
\BIBentrySTDinterwordspacing

\bibitem{schulman2017proximal}
J.~Schulman, F.~Wolski, P.~Dhariwal, A.~Radford, and O.~Klimov, ``Proximal policy optimization algorithms,'' \emph{arXiv preprint arXiv:1707.06347}, 2017.

\end{thebibliography}

\end{document}